\documentclass{article}

\usepackage[final,nonatbib]{neurips_2019}

\usepackage{hyperref}
\usepackage{url}
\usepackage{amsfonts}       
\usepackage{nicefrac}       
\usepackage{microtype}      

\usepackage{paralist}
\usepackage{multirow}
\usepackage{graphicx}
\usepackage[utf8]{inputenc} 
\usepackage[T1]{fontenc}    
\usepackage{hyperref}       
\usepackage{url}            
\usepackage{booktabs}       
\usepackage{amsfonts}       
\usepackage{nicefrac}       
\usepackage{microtype}      
\usepackage{amsmath}
\usepackage{color}
\usepackage{amsmath}
\usepackage[ruled,vlined]{algorithm2e}
\usepackage{graphicx}
\usepackage{subfig}
\usepackage{color}
\usepackage{listings} 
\usepackage{xcolor}
\usepackage{floatrow}
\usepackage{enumitem}
\usepackage{siunitx}
\title{CrevNet: Conditionally Reversible Video Prediction}

\author{Wei Yu, Yichao Lu, Steve Easterbrook, Sanja Fidler\\
University of Toronto\\
{\tt\small \{gnosis,yichao,sme,fidler\}@cs.toronto.edu}
}

\begin{document}

\maketitle

\begin{abstract}
Applying resolution-preserving blocks is a common practice to maximize information preservation in video prediction, yet their high memory consumption greatly limits their application scenarios.
We propose CrevNet, a Conditionally Reversible Network that uses reversible architectures to build a bijective two-way autoencoder and its complementary recurrent predictor. Our model enjoys the theoretically guaranteed property of no information loss during the feature extraction, much lower memory consumption and computational efficiency. 
\end{abstract}

\section{Introduction}
Most of the existing models for video prediction employ a hybrid of convolutional and recurrent layers as the underlying architecture~(\cite{wang2017predrnn,xingjian2015convolutional,lotter2016deep}). Such architectural design enables the model to simultaneously exploit the ability of convolutional units to model spatial relationships and the potential of recurrent units to capture temporal dependencies. Despite their prevalence in the literature, classical video prediction architectures suffer from one major limitation. In dense prediction tasks such as video prediction, models are required to make pixel-wise predictions, which emphasizes the demand for the preservation of information through layers. Prior works attempt to address such demand through the extensive use of resolution-preserving blocks~(\cite{wang2017predrnn,wang2018predrnn++,kalchbrenner2016video}). Nevertheless, these resolution-preserving blocks are not guaranteed to preserve all the relevant information, and they greatly increase the memory consumption and computational cost of the models.

Recently, reversible architectures~(\cite{dinh2014nice,gomez2017reversible,jacobsen2018revnet}) have attracted attention due to their light memory demand and their information preserving property by design. However, the effectiveness of reversible models remains greatly unexplored in the video literature. In this extended abstract, we introduce a novel, conditionally reversible video prediction model, CrevNet,  in the sense that when conditioned on previous hidden states, it can exactly reconstruct the input from its predictions. The contribution of this work can be summarized as follows:
\begin{itemize}
\item We introduce a two-way autoencoder that uses the forward and backward passes of an invertible network as encoder and decoder. The volume-preserving two-way autoencoder not only greatly reduces the memory demand and computational cost, but also enjoys the theoretically guaranteed property of no information loss. 
\item We propose the reversible predictive module (RPM), which extends the reversibility from spatial to temporal domain. RPM, together with the two-way autoencoder, provides a conditionally reversible architecture (CrevNet) for spatiotemporal learning. 
\end{itemize}

\section{Approach}
\label{section:Approach}
We first outline the general pipeline of our method. Our CrevNet consists of two subnetworks, an autonencoder network with an encoder $\mathcal{E}$,  decoder $\mathcal{D}$ and a recurrent predictor $\mathcal{P}$ bridging encoder and decoder. Let $x_t\in \mathbb{R}^{w\times h\times c}$ represent the $t_{\text{th}}$ frame in video $x$, where $w$, $h$, and $c$ denote its width, height, and the number of channels. Given $x_{0:t-1}$, the model predicts the next frame $\hat{x}_t$ as follows:
\begin{equation}
    \hat{x}_t = \mathcal{D}(\mathcal{P}(\mathcal{E}(x_{t-1})|x_{0:t-2}))
    \end{equation}
During the multi-frame generation process without access to the ground truth frames, the model uses its previous predictions instead.

\begin{figure*}
	\centering
		\includegraphics[width=0.9\textwidth]{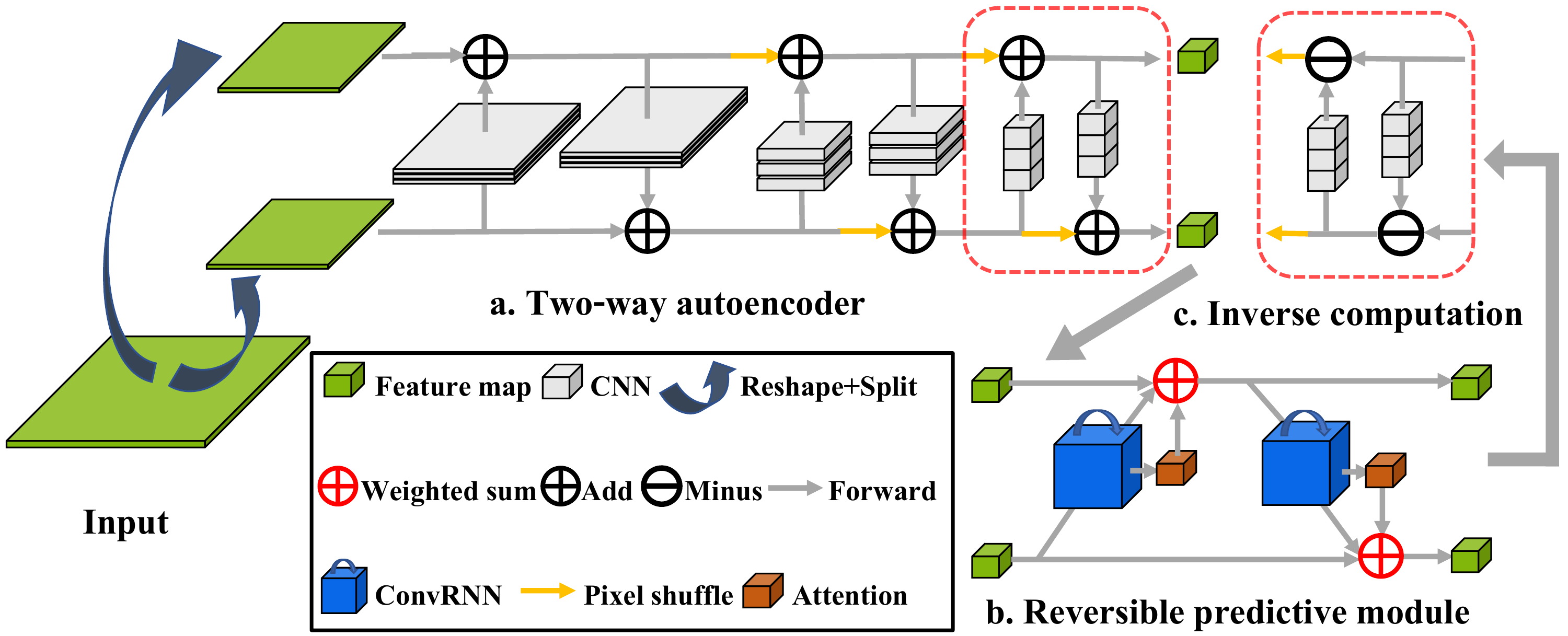}
		\caption{The network architecture of CrevNet (Better viewed in color). The input video frames are first reshaped and split channelwise into two groups. These two groups are passed to the two-way autoencoder (a) for feature extraction, and then to the predictor made up of multiple reversible predictive modules (b). The transformed high-level features produced by predictor are then passed back through the decoding pass of (a), shown here as a representative block (c) to yield its prediction.}
		\vspace{-2pt}
		\label{figure:model}
\end{figure*}

\subsection{The Invertible Two-way Autoencoder}
We propose a bijective two-way autoencoder based on the additive coupling layer introduced in NICE~(\cite{dinh2014nice}). We begin with describing the building block of the two-way autoencoder (Fig~\ref{figure:model}{\color{red} a}). Formally, the input $x$ is first reshaped and split channelwise into two groups, denoted as $x^1$ and $x^2$. During the forward pass of each building block, one group, e.g. $x^1$, passes through several convolutions and activations and is then added to another group, $x^2$, like a residual block:
 \begin{align}
 \hat{x} ^2= x^2+\mathcal{F}_1(x^1)  ~~~~~~~~~~~~~~~~   \hat{x}^1 = x^1+\mathcal{F}_2(\hat{x}^2)  
\end{align}
where $\mathcal{F}$ is a composite non-linear operator consisting of convolutions and activations, and $\hat{x}^1$ and $\hat{x}^2$ are the updated $x^1$ and $x^2$. Note that $x^1$ and $x^2$ can be simply recovered from $\hat{x}^2$ and $\hat{x}^1$ by the inverse computation (Fig~\ref{figure:model}{\color{red} c}) as follows:
 \begin{align}
 x^1 = \hat{x}^1 -\mathcal{F}_2(\hat{x}^2) ~~~~~~~~~~~~~~~~  x^2 = \hat{x}^2 -\mathcal{F}_1(x^1)
\end{align}
 Multiple building blocks are stacked in an alternating fashion between $x^1$ and $x^2$ to construct a two-way autoencoder, as shown in Fig~\ref{figure:model}{\color{red} a}. A series of the forward and inverse computations builds a one-to-one and onto, i.e. bijective , mapping between the input and features. Such invertibility ensures that there is no information loss during the feature extraction, which is presumably more favorable for video prediction since the model is expected to restore the future frames with fine-grained details. To enable the invertibility of the entire autoencoder, our two-way autoencoder uses a bijective downsampling, pixel shuffle layer~(\cite{shi2016real}), that changes the shape of feature from $(w, h, c)$ to $(w/n, h/n, c \times n^2)$. The resulting volume-preserving architecture can greatly reduce its memory consumption compared with the existing resolution-preserving methods.

We further argue that for generative tasks, e.g. video prediction, we can effectively utilize a single two-way autoencoder, and to use its forward and backward pass as the encoder and the decoder, respectively. The predicted frame $\hat{x}_{t}$ is thus given by
\begin{equation}
\hat{x}_{t} = \mathcal{E}^{-1}(\mathcal{P}(\mathcal{E}(x_{t-1})|x_{0:t-2})) 
\end{equation}
where $\mathcal{E}^{-1}$ is the backward pass of $\mathcal{E}$. Our rationale is that, such setting would not only reduce the number of parameters in the model, but also encourage the model to explore the shared feature space between the inputs and the targets. As a result, our method does not require any form of information sharing, e.g. skip connection, between the encoder and decoder. In addition, our two-way autoencoder can enjoy a lower computational cost at the multi-frame prediction phase where the encoding pass is no longer needed and the predictor directly takes the output from previous timestep as input, since $\mathcal{E}(\mathcal{E}^{-1})$ is an identity mapping .

\subsection{Reversible Predictive Module}

In this section, we describe the second part of our video prediction model, the predictor $\mathcal{P}$, which computes dependencies along both the space and time dimensions. Although the traditional stacked-ConvRNN layers architecture is the most straightforward  choice of predictor, we find that it fails to establish a consistent temporal dependency  when equipped with our two-way autoencoder through experiments. Therefore, we propose a novel reversible predictive module (RPM), which can be regarded as a recurrent extension of the two-way autoencoder. In the RPM, we substitute all standard convolutions with layers from the ConvRNN family (e.g. ConvLSTM or spatiotemporal LSTM) and introduce a soft attention (weighting gates) mechanism to form a weighted sum of the two groups instead of the direct addition. The main operations of RPM used in this paper are given as follows:
\begin{align*}
h_t^1 &=\text{ConvRNN}(x_t^1,h_{t-1}^1)  && \textit{ConvRNN} \\
g_t &=\phi (W_2*\text{ReLU}(W_1*h_t^1+ b_1)+b_2)  && \textit{Attention module}\\
\hat{x_t^2} &= (1-g_t)\odot x_t^2+g_t\odot h_t^1  && \textit{Weighted sum}
\end{align*}
where $x_t^1$ and $x_t^2$ denote two groups of features at timestep $t$, $h_t^1$ denote the hidden states of ConvRNN layer, $\phi$ is sigmoid activation, $*$ is the standard convolution operator and $\odot$ is the Hadamard product.  The architecture of reversible predictive module is also shown in Fig~\ref{figure:model}{\color{red} b}. RPM adopts a similar architectural design as the two-way autoencoder to ensure a pixel-wise alignment between the input and the output, i.e. each position of features can be traced back to certain pixel, and thus make it compatible with our two-way autoencoder. It also mitigates the vanishing gradient issues across stacked layers since the coupling layer provides a nice property w.r.t. the Jacobian~(\cite{dinh2014nice}).  In addition, the attention mechanism in the RPM enables the model to focus on objects in motion instead of background, which further improves the video prediction quality. 
Similarly, multiple RPMs alternate between the two groups to form a predictor. We call this predictor conditionally reversible since, given $h_{t-1}$, we are able to reconstruct $x_{t-1}$ from $\hat{x}_t$  if there are no numerical errors:
\begin{equation}
{x}_{t-1} = \mathcal{E}^{-1}(\mathcal{P}^{-1}(\mathcal{E}(\hat{x}_{t})|h_{t-1}))
\end{equation}
where $\mathcal{P}^{-1}$ is the inverse computation of the predictor $\mathcal{P}$. We name the video prediction model using two-way autoencoder as its backbone and RPMs as its predictor CrevNet. Another key factor of RPM is the choice of ConvRNN. In this paper, we mainly employ ConvLSTM~(\cite{xingjian2015convolutional}) and spatiotemporal LSTM (ST-LSTM,~\cite{wang2017predrnn}). 

\section{Experiment}

We evaluate our model on a more complicated real-world dataset, Traffic4cast, which collects the traffic statuses of 3 big cities over a year at a 5-minute interval. Traffic forecasting can be straightforwardly defined as video prediction task by its spatiotemporal nature. However, this dataset is quite challenging for the following reasons. (1). High resolution: The frame resolution of Traffic4cast is 495 $\times$ 436, which is the highest among all datasets. Existing resolution-preserving methods can hardly be adapted to this dataset since they all require extremely large memory and computation. Even if these models can be fitted in GPUs, they still do not have large enough receptive fields to capture the meaningful dynamics as vehicles can move up to 100 pixels between consecutive frames. (2). Complicated nonlinear dynamics: Valid data points only reside on the hidden roadmap of each city, which is not explicitly provided in this dataset.  Moving vehicles on these curved roads along with tangled road conditions will produce very complex nonlinear behaviours. It also involves many unobservable conditions or random events like weather and car accidents.

\textbf{Datasets and Setup:} Each frame in Traffic4cast dataset is a 495 $\times$ 436 $\times$ 3 heatmap, where the last dimension records 3 traffic statuses representing volume, mean speed and major direction at given location. The general architecture of CrevNet is composed of a 36-layer two-way autoencoder and 8 RPMs. All variants of CrevNet are trained by using the Adam optimizer with a starting learning rate of \num{2e-4} to minimize MSE. We train each model to predict next 3 frames (the next 15 minutes) from 9 observations and evaluate prediction with MSE criterion. Moreover, we found another two tricks , finetuning on each city and finetuning on different timestep , that could further improve the overall results.

	\begin{table}[htb]
	\small
		\begin{tabular}{ p{1.9cm}|p{1.8cm}|p{2.9cm}|p{3.4cm}|p{1.8cm}  }
			\toprule
			CrevNet  &single model &  finetuning on each city &  finetuning on each timestep & final model \\
	
			\midrule
			
			Per-pixel MSE  & \num{9.392e-3} & \num{9.340e-3} &\num{9.285e-3} &\num{9.251e-3}\\
			\bottomrule
		\end{tabular}
		\caption{Quantitative evaluation on Traffic4cast. Lower MSE indicates better prediction accuracy.}  \label{tab:traffic}

	\end{table}


\textbf{Results:} The quantitative comparison is provided in Table~\ref{tab:traffic}. Unlike all previous state-of-the-art methods, CrevNet does not suffer from high memory consumption so that we were able to train our model in a single V100 GPU. The invertibility of two-way autoencoder preserves all necessary information for spatio-temporal learning and 
allows our model to generate sharp and reasonable predictions. 

\bibliography{main}
\end{document}